\documentclass[final,3p,times,authoryear]{elsarticle}
\usepackage[T1]{fontenc}
\usepackage{iftex}
\ifPDFTeX
  \usepackage[utf8]{inputenc}
\fi
\usepackage{amsmath,amssymb,booktabs,graphicx,adjustbox,multirow,makecell}
\usepackage{array,tabularx,microtype}
\usepackage{xurl}
\usepackage[hidelinks]{hyperref}
\biboptions{sort&compress}
\newcommand{\authorquery}[1]{}

\journal{Computer Vision and Image Understanding}
\begin{document}
\begin{frontmatter}
\title{From Retouched Images to Paired Videos: A Synthetic Benchmark for Face Video Retouching}
\author[aff1,aff2]{Peiyuan Zeng}
\ead{andyzeng0921@outlook.com}
\author[aff1]{Fengchuang Xing\corref{cor1}}
\ead{xingfengchuang@gdei.edu.cn}
\author[aff1]{Shiming Wu}
\ead{hotgirllllllll@outlook.com}
\author[aff1]{Hongwei Yue}
\ead{springfall@foxmail.com}
\author[aff3]{Chao Zhou}
\ead{zhouchaodzkd@e.gzhu.edu.cn}
\cortext[cor1]{Corresponding author.}
\affiliation[aff1]{organization={College of Physics and Information Engineering,
  Guangdong University of Education},country={China}}
\affiliation[aff2]{organization={Cloudbutterfly Technology}}
\affiliation[aff3]{organization={School of Electronic Information, Huzhou College},
  city={Huzhou},postcode={313000},country={China}}

\begin{abstract}
Paired unretouched/retouched face videos are costly to produce because professional editing must remain consistent over time, while releasing identifiable face footage raises privacy, consent, and governance concerns. We present DeepVFR, an initial controlled synthetic benchmark---not a new retouching architecture---for paired face video retouching. Starting from paired synthetic DeepFRR images, Wan2.2 generates mild-motion source clips, and backend-specific face transfer constructs synchronized proxy retouching targets, followed by automatic validity screening and manual visual inspection. DeepVFR contains 1,024 pairs of 150-frame clips at 30 fps, evaluated under an identity-disjoint protocol with explicit frame correspondence; release metadata will record the construction route. Within-benchmark baselines, external checkpoints applied to DeepVFR, qualitative analysis, and a 17-participant ranking study establish reference behavior in the controlled domain. The constructed targets exhibit a substantial low-preference tail in the human study. Because both streams inherit synthetic content and the targets additionally inherit commercial-API or face-transfer artifacts, the benchmark neither replaces artist-retouched captured video nor establishes synthetic-to-real generalization. DeepVFR provides a starting point for paired supervision and controlled evaluation. Data, individual study rankings, and training logs will be made publicly available upon acceptance. Broader motion coverage and real-video validation remain future work.

\end{abstract}
\end{frontmatter}
\section{Introduction}
Portrait videos have become a central form of visual self-presentation on social media, turning face retouching from a photo-editing operation into a large-scale video-processing problem. With the rapid growth of short-form video platforms and mobile content creation~\citep{datareportal2025digital}, users increasingly expect face-centric videos to be enhanced in terms of skin texture, facial tone, blemish removal, and overall portrait quality. However, manual video retouching remains labor-intensive and difficult to scale, since edits must be applied consistently across frames while preserving identity, expression, pose, and natural facial motion.

Automatic face and portrait retouching has been extensively studied in the image domain. General and portrait photo retouching datasets, such as MIT-Adobe FiveK~\citep{bychkovsky2011learning} and PPR10K~\citep{liang2021ppr10k}, have enabled learning-based photographic enhancement and portrait-specific retouching. Face-oriented resources cover a broader range of retouching-related settings. EFrR studies exemplar-based freckle retouching using before-and-after facial images from cosmetic laser treatment~\citep{lin2019exemplar}. FFHQ provides high-quality face images for generative modeling~\citep{Karras2018StyleGAN}, and FFHQR builds upon it with professionally retouched face images for automatic face retouching~\citep{shafaei2021autoretouch}. RetouchingFFHQ further scales face retouching data to over half a million retouched face images for fine-grained retouching detection~\citep{ying2023retouchingffhq}. DigiFace-1M provides more than one million synthetic face images with diverse poses, lighting, and backgrounds, mainly for face recognition~\citep{bae2023digiface}. More recently, HGFR compares these representative resources, including FRTDD, FFHQ, and FFHQR, and introduces 25,000 high-definition paired face retouching images with rich blemish types~\citep{xu2025face}.

Despite this growing dataset landscape, existing resources remain fundamentally image-centric. They focus on portrait photo retouching, face generation, face recognition, retouching detection, or paired image-level blemish removal, but do not provide temporally aligned before-and-after face videos. This limitation is critical for video face retouching. Applying an image retouching model frame by frame often causes temporal flickering, unstable facial details, and inconsistent identity-related attributes, a problem also observed in general video processing and portrait video stylization tasks~\citep{lai2018learning,yang2022vtoonify}. Therefore, learning a practical video face retouching model requires paired video-level supervision, where the input and retouched videos share the same identity, expression, pose, motion, and camera trajectory while differing mainly in retouching style.

Constructing such paired video data in the real world is difficult. Producing high-quality retouched videos requires professional editing and careful temporal adjustment, making large-scale annotation expensive. Moreover, face videos contain sensitive identity information, and collecting paired unretouched/retouched videos raises additional privacy, consent, and dataset governance concerns~\citep{gebru2021datasheets}. As a result, although image-level face retouching datasets have become increasingly available, there is still no widely adopted paired benchmark for video face retouching.

The most related supervision sources are FFHQR-Seq and MRFV introduced in VRetouchEr~\citep{Xue2024VRetouchEr}. As summarized in Tab.~\ref{tab:benchmark_comparison}, FFHQR-Seq is constructed from FFHQR image pairs by applying random cropping, flipping, and translation to generate short paired sequences. Although it provides many paired samples, its temporal variations mainly come from geometric augmentation rather than facial motion, expression dynamics, or camera movement. MRFV contains 200 in-the-wild portrait videos with manually retouched frames and is therefore closer to the intended real-video task, but its construction is costly and the data have not been publicly released in the official repository. Existing resources consequently leave a practical gap for a fixed, inspectable benchmark with seconds-long paired clips.

To address this gap, we present DeepVFR (video face retouching), distinct from its image source DeepFRR, an initial controlled synthetic benchmark that converts paired image supervision into paired video supervision. Starting from each raw/retouched image pair, an image-to-video generator supplies a shared mild-motion trajectory, after which backend-specific face transfer constructs a synchronized proxy retouching target. The off-the-shelf generators are construction tools rather than a proposed retouching architecture. Our contribution is the paired benchmark, its supervision interface, and a unified evaluation protocol. The resulting targets are not professionally edited captured videos, and we therefore restrict our claims to the controlled synthetic domain.
Our contributions are summarized as follows:
\begin{itemize}
    \item We introduce an initial controlled paired-video benchmark for face retouching. It contains 1,024 raw and constructed-target clip pairs, each with 150 frame correspondences, and uses identity-disjoint splits. The 1,024 clips---rather than the correlated frame count---are the benchmark's clip-level temporal units.
    
    \item We specify a benchmark-construction and planned-release protocol that combines image-to-video synthesis, Akool-based primary transfer, a smaller complementary SimSwap route, quality control, per-clip route metadata, and explicit raw/target frame correspondence. This protocol supplies paired supervision without claiming a new generation model.
    
    \item We report controlled-domain baselines using frame-fidelity and temporal metrics, released external checkpoints, qualitative analysis, and a 17-participant ranking study. The experiments establish reference behavior on DeepVFR; they do not establish transfer from DeepVFR to artist-retouched or in-the-wild captured videos.

\end{itemize}

\begin{table}[t]
    \centering
    \small
    \caption{Representative supervision sources for video face retouching.}
    \label{tab:benchmark_comparison}
    \setlength{\tabcolsep}{4pt}
    \begin{tabularx}{\linewidth}{@{}>{\raggedright\arraybackslash}p{.19\linewidth}>{\raggedright\arraybackslash}p{.14\linewidth}>{\raggedright\arraybackslash}p{.15\linewidth}>{\raggedright\arraybackslash}Xcc@{}}
        \hline
        Benchmark & Scale & Temporal unit & Motion source & \makecell{Public\\split} & \makecell{Paired\\beauty task} \\
        \hline
        FFHQR-Seq~\citep{Xue2024VRetouchEr} & 56k/7k/7k seq. & 5 frames/sample & Geometric augmentation & Yes & Yes \\
        MRFV~\citep{Xue2024VRetouchEr} & 200 videos & $\geq$500 frames/video & Real portrait videos & No & Yes \\
        DeepVFR (ours) & 1,024 videos & 150 frames/video & Paired I2V videos & Planned & Yes \\
        \hline
    \end{tabularx}
    
\end{table}
 
\section{Related Work}\label{sec:related}

\noindent\textbf{Image face retouching datasets.}
Image-level retouching predates video-level supervision and has been studied extensively. Representative image-level resources related to facial retouching are summarized in Tab.~\ref{tab:image_beauty_datasets}. General photo-retouching datasets such as MIT-Adobe FiveK~\citep{bychkovsky2011learning} and portrait-specific datasets such as PPR10K~\citep{liang2021ppr10k} provide paired image supervision, but they mainly focus on global photographic enhancement or portrait tone adjustment rather than face-level retouching. Face-oriented resources cover beauty perception, makeup transfer, and face retouching detection. SCUT-FBP and SCUT-FBP5500 provide facial attractiveness annotations~\citep{xie2015scut,liang2018scut}; EFrR studies freckle removal and skin-tone adjustment using before/after cosmetic-treatment exemplars~\citep{lin2019exemplar}; FFHQ and FFHQR provide high-quality face images and retouched face images~\citep{Karras2018StyleGAN,shafaei2021autoretouch}; RetouchingFFHQ scales fine-grained face retouching detection to over half a million conditionally retouched images~\citep{ying2023retouchingffhq}. More recently, HGFR summarizes image-level face retouching resources and introduces 25,000 high-definition paired face retouching images~\citep{xu2025face}. These datasets are valuable for image retouching, but they do not provide frame-synchronized raw/retouched video pairs.

Among these resources, we choose DeepFRR as the image source for DeepVFR for three reasons~\citep{FRRffusion2024}. First, DeepFRR provides one-to-one raw/retouched face pairs, which is necessary for constructing paired supervision. Second, its images are high-resolution synthetic faces generated by StyleGAN, reducing the privacy risk compared with directly collected real identities~\citep{StyleGAN22020}. Third, its 50k-pair scale gives a sufficiently large candidate pool for quality filtering, demographic balancing, and replacement of failed video-generation or transfer cases. Therefore, DeepFRR is more suitable than unpaired beauty-score datasets, detection-oriented retouching datasets, or general portrait-retouching datasets for constructing paired face video retouching supervision.

\noindent\textbf{Face swapping and identity transfer.}
Face swapping aims to transfer the identity or facial appearance of a source face to a target face while preserving target pose, expression, illumination, and background. Early and representative studies include automatic face reenactment~\citep{Garrido2014Reenactment}, Face2Face~\citep{Thies2016Face2Face}, X2Face~\citep{Wiles2018X2Face}, Deep Video Portraits~\citep{Kim2018DeepVideoPortraits}, FSGAN~\citep{Nirkin2019FSGAN}, and Neural Textures~\citep{Thies2019NeuralTextures}. Later approaches include FaceShifter~\citep{Li2020FaceShifter}, SimSwap~\citep{SimSwap2021}, MegaFS~\citep{Zhu2021MegaFS}, HifiFace~\citep{Wang2021HifiFace}, high-resolution latent face swapping~\citep{Xu2022HighResolutionFaceSwapping}, FaceDancer~\citep{Rosberg2023FaceDancer}, 3D-aware face swapping~\citep{Li2023ThreeDAwareFaceSwapping}, and BlendFace~\citep{Shiohara2023BlendFace}. In addition to academic models, commercial APIs such as the Akool Face Swap API provide face-transfer interfaces for image and video processing~\citep{AkoolAPI2025}. In this work, face swapping is a construction tool for transferring the paired retouched appearance, not the retouching model being benchmarked.

\noindent\textbf{Video face retouching datasets.}
Compared with image face retouching, video face retouching requires temporally aligned supervision. Existing face video datasets such as FaceForensics++~\citep{FaceForensicsPP2019} and CelebV-HQ~\citep{CelebVHQ2022} contain facial videos, but they are not designed for paired raw and retouched video supervision. The most related resources are FFHQR-Seq and MRFV introduced in VRetouchEr~\citep{Xue2024VRetouchEr}. FFHQR-Seq is built from FFHQR image pairs by random cropping, flipping, and translation, producing short pseudo-sequences rather than natural facial motion. MRFV contains 200 manually retouched portrait videos, but it is small and not publicly released in the official repository~\citep{VRetouchErRepo2024}. These limitations motivate DeepVFR, which converts paired image supervision into seconds-long, frame-synchronized video pairs.

Recent model-centric work explores mixtures of diffusion models, instruction-following retouching agents, and reward-guided aesthetic alignment~\citep{Liu2025MoFRR,Wu2026RetouchIQ,Yang2026BeautyGRPO}. These methods broaden the space of retouching operators, whereas DeepVFR addresses the complementary problem of paired temporal supervision and controlled evaluation. Deepfake resources such as DF40 and FakeAVCeleb focus on identity manipulation or forgery analysis~\citep{Yan2024DF40,Khalid2021FakeAVCeleb}. Their real/fake labels and identity-replacement targets do not provide the same-identity raw/beauty correspondence required for supervised face video retouching.

\begin{table}[t]
    \centering
    \caption{Representative image-level resources related to facial beauty and retouching. ``Paired'' indicates whether an original/beauty or before/after image relation is available.}
    \label{tab:image_beauty_datasets}
    \small\setlength{\tabcolsep}{4pt}
    \begin{tabularx}{\linewidth}{@{}>{\raggedright\arraybackslash}p{.35\linewidth}>{\raggedright\arraybackslash}p{.18\linewidth}cX@{}}
        \hline
        Dataset & Scale & Paired & Main focus \\
        \hline
        MIT-Adobe FiveK~\citep{bychkovsky2011learning} & 5k images & Yes & General photo retouching \\
        PPR10K~\citep{liang2021ppr10k} & 11,161 portraits & Yes & Portrait photo retouching \\
        SCUT-FBP5500~\citep{liang2018scut} & 5,500 faces & No & Beauty perception \\
        EFrR~\citep{lin2019exemplar} & Small exemplar set & Yes & Freckle/skin retouching \\
        FFHQ/FFHQR~\citep{Karras2018StyleGAN,shafaei2021autoretouch} & Face images & Partial & Face generation/retouching \\
        RetouchingFFHQ~\citep{ying2023retouchingffhq} & $>$0.5M images & Conditional & Retouching detection \\
        DeepFRR~\citep{FRRffusion2024} & 50k pairs & Yes & Face retouching reversal \\
        HGFR~\citep{xu2025face} & 25k pairs & Yes & High-definition face retouching \\
        \hline
    \end{tabularx}
    
\end{table}

\begin{figure}[t]
    \centering
    \includegraphics[width=\textwidth]{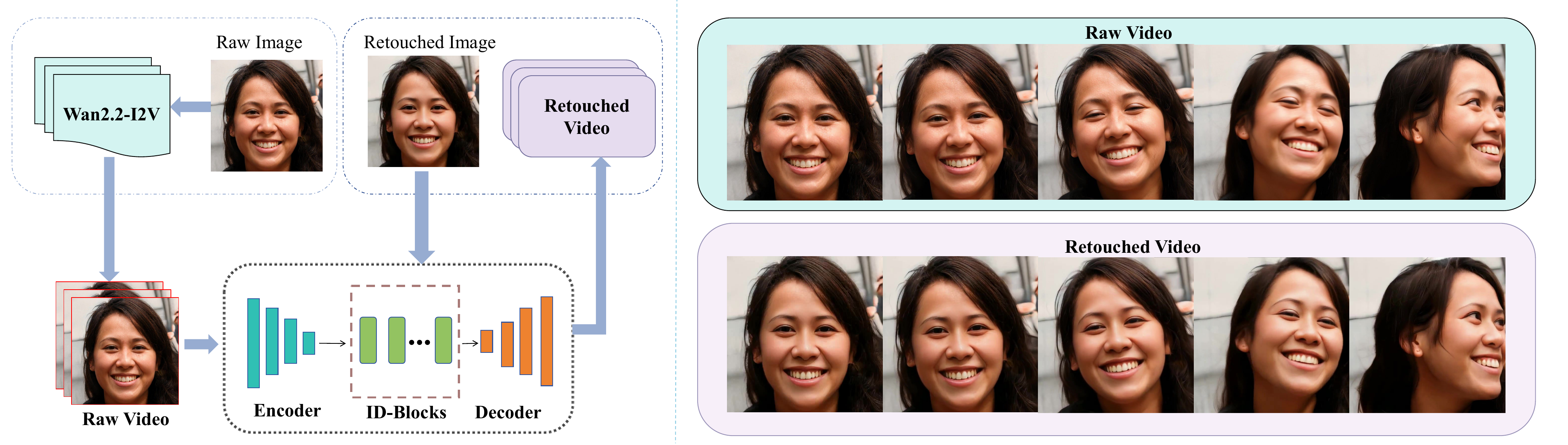}
    \caption{Overview of the DeepVFR construction pipeline. An I2V model animates the raw member of each DeepFRR pair; backend-specific transfer and parsing-guided blending then construct a frame-synchronized proxy target from the paired retouched image. The encoder/ID-block/decoder schematic illustrates the complementary SimSwap route; it does not describe the proprietary Akool backend. The right side shows representative raw/target frames.}
    \label{fig:i2v_pipeline}
\end{figure}

\section{Benchmark Construction and Protocol}\label{sec:method}
This section specifies how the DeepVFR collection is constructed and how its paired supervision is used. It does not introduce a new retouching network. We describe source-pair curation, I2V-based motion synthesis, backend-specific target construction, quality control, and the common training/evaluation interface; release-specific identifiers and integrity records will accompany the data package planned for release upon acceptance.
\subsection{Overview of the DeepVFR Construction Pipeline}
We construct DeepVFR through a paired image-to-video pipeline, as illustrated in Fig.~\ref{fig:i2v_pipeline}. Given a raw/retouched image pair from DeepFRR, we first generate a raw video from the raw image and then construct a temporally aligned proxy target using the paired retouched image. Specifically, the raw face image is animated into a short portrait video, after which the retouched appearance is transferred frame by frame while attempting to retain the generated pose, expression, and motion trajectory. This produces synchronized pairs for training and evaluating face video retouching models under controlled paired supervision.


\subsection{Source Pair Curation from DeepFRR}

Let $D_{\mathrm{img}} = \{(x_i, y_i^b)\}_{i=1}^{N}$ denote the DeepFRR image corpus, where $x_i$ is the raw face image and $y_i^b$ is its corresponding retouched image~\citep{FRRffusion2024}.
In the actual construction process, we select 1,024 image pairs from DeepFRR through a combination of quality filtering and stratified sampling. The filtering procedure follows four criteria. First, each sample pair must contain a single clear frontal or near-frontal face in both the raw image and the retouched image to ensure stable alignment and transfer. Second, the facial region should have sufficient resolution, complete facial components, and no severe occlusion, watermark, or obvious generation artifacts. Third, the retouched image should preserve identity while exhibiting visible but not excessive retouching effects, so as to avoid style distortion in the constructed results. Fourth, the selected subset should cover diverse ages, genders, skin tones, illumination conditions, and expression tendencies as much as possible, thereby improving the overall diversity of the dataset distribution.

If a sample fails a later generation or transfer check, it is replaced with the next valid candidate from the same stratum. This procedure yields 1,024 retained pairs that satisfy the stated inclusion criteria. It does not make the groups perfectly balanced; the remaining demographic imbalance is reported in Sec.~\ref{sec:dataset}.

\subsection{I2V-Based Raw Video Synthesis}

For each selected raw image $x_i$, we generate a corresponding raw video using the Tongyi Wanxiang 2.2 image-to-video model ~\citep{wan2025,Wan2Repo2025,TongyiWanxiang22I2VFlash}. It should be emphasized that the I2V model is used here only as a motion-prior generator rather than as the final retouching model. To induce natural micro-expressions and slight head motion while avoiding large pose changes that may break identity alignment, we adopt a mild expression prompt, such as \texttt{smile}, during video generation. For each sample, the generated video is uniformly set to $T = 150$ frames with a frame rate of $f = 30$ fps, corresponding to a clip of approximately 5 seconds:

\begin{equation}
O_i = \{O_{i,t}\}_{t=1}^{T} = G_{\mathrm{I2V}}(x_i; p, T, f), \quad T = 150, \quad f = 30,
\end{equation}
where $G_{\mathrm{I2V}}$ denotes the image-to-video generator, $p$ denotes the text prompt, and $O_{i,t}$ denotes the raw video generated in frame $t$.

We adopt this setting for three reasons. First, 150-frame clips expose adjacent-frame flicker and appearance drift over a longer span than five-frame pseudo-sequences. Second, 30 fps is compatible with common video-processing pipelines. Third, about 5 seconds balances temporal coverage, generation failures, storage, and training cost. The resulting trajectories are deliberately mild; they are useful for controlled temporal evaluation but do not cover rapid speech, large pose changes, occlusion, or dynamic lighting.

\subsection{Face-Transfer-Based Retouched Video Construction}

After obtaining the raw video $O_i$ and its corresponding retouched image $y_i^b$, we construct the proxy retouched video $B_i = \{B_{i,t}\}_{t=1}^{T}$. For each $O_{i,t}$, we use $y_i^b$ as the source appearance and the current raw frame as the motion/background target, then apply a backend-specific face-transfer operator $T_{b_i}$, where $b_i$ denotes SimSwap or the Akool API. The process is

\begin{equation}
\tilde{B}_{i,t} = T_{b_i}\!\left(\mathcal{A}_{b_i}(y_i^b, O_{i,t})\right),
\end{equation}
\begin{equation}
B_{i,t} = \alpha(M_{i,t}) \odot \tilde{B}_{i,t} + \left(1 - \alpha(M_{i,t})\right) \odot O_{i,t},
\end{equation}
where $\mathcal{A}_{b_i}$ denotes the detector-and-alignment transform required by backend $b_i$, implemented with SCRFD or RetinaFace~\citep{SCRFD2021,Deng2020RetinaFace}; $M_{i,t}$ is the BiSeNet face-parsing mask~\citep{Yu2018BiSeNet}; $\alpha(M_{i,t})$ is its feathered alpha mask; and $\odot$ is element-wise multiplication. The blend retains the background, hair, and non-face regions of $O_{i,t}$ while inserting the transferred facial region.

Akool is the primary production transfer route, while SimSwap supplies a smaller complementary subset~\citep{SimSwap2021,AkoolAPI2025}. Under SimSwap, identity features from the paired retouched image condition the generator while the target frame supplies pose and expression. Identity conditioning does not explicitly encode skin-retouching edits, so the transferred appearance remains proxy supervision. Both routes are converted to the same clip format, and the planned release manifest will identify the backend used for each clip. Because Akool is a mutable commercial service, a future versioned fixed package and its provenance could support repeatable scoring, but exact regeneration may change when the service changes.

After transfer, automatic checks verify detection and alignment, while visual inspection rejects severe motion collapse, obvious boundary artifacts, loss of the intended retouched appearance, or clear identity drift. Rejected clips are replaced from the same sampling stratum. This hybrid screen removes conspicuous failures from the current 1,024 pairs; it cannot detect every latent synthetic or backend artifact, and its manual component is a bottleneck for expansion.

\subsection{Paired Supervision Protocols Supported by DeepVFR}
After the above steps, the resulting DeepVFR dataset can be expressed as
\begin{equation}
D_{\mathrm{VFR}} = \{(O_i, B_i)\}_{i=1}^{1024},
\end{equation}
where $O_i$ denotes the raw video and $B_i$ denotes the frame-aligned retouched video. Each sample contains 150 synchronized frames, so that every raw frame $O_{i,t}$ has a corresponding retouched target $B_{i,t}$. This representation makes DeepVFR directly usable under both image-level and video-level retouching settings.

For image-based retouching methods, DeepVFR can be decomposed into individual frame pairs $(O_{i,t}, B_{i,t})$, allowing existing image models to be trained without changing their input--output formulation. For video-based methods, the ordered sequences $(O_i, B_i)$ share frame indices and are constructed to follow the same nominal motion and expression trajectory, enabling models to exploit cross-frame information rather than relying only on pseudo-sequences or frame-wise augmentation.

We keep this protocol intentionally simple. All methods take raw frames or clips as input and use the corresponding retouched frames or clips as supervision, with the sampling unit being the only difference between single-frame and temporal-sequence settings. In this way, DeepVFR separates the benchmark contribution from any particular network design and provides a controlled basis for comparing image-based and video-based face retouching methods under the same paired supervision source.

 \begin{figure}[t]
        \centering
        \includegraphics[width=\textwidth]{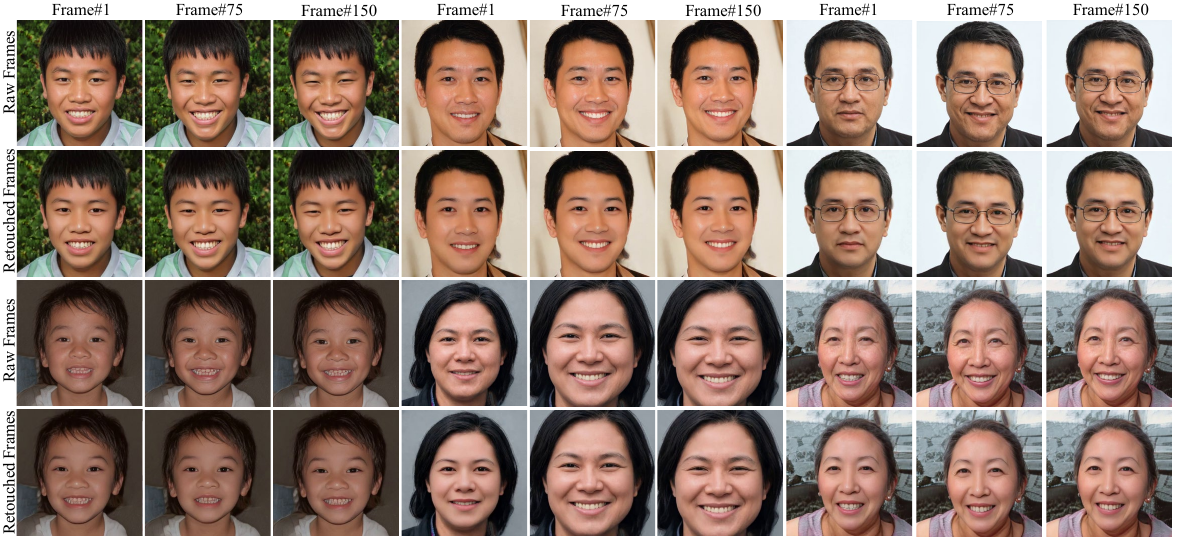}
        \caption{Representative DeepVFR samples. Each group contains three frames from one clip; the six examples were selected from metadata strata crossing age group (child, youth, elderly) and gender (male, female).}
        \label{fig:dataset_overview}
\end{figure}

\section{DeepVFR Dataset}\label{sec:dataset}
\subsection{Dataset Contents and Scope}

DeepVFR is a controlled synthetic benchmark for video face retouching. Each sample contains a generated raw portrait clip and a frame-synchronized proxy retouching target constructed from the same nominal trajectory. The streams therefore share frame indices and are constructed to follow corresponding pose, expression, and motion, while differing primarily in transferred facial appearance. VFR in ``DeepVFR'' stands for video face retouching; the name refers to the benchmark task rather than to a new model.

The dataset contains 1,024 paired clips. Each clip has 150 frames at 30 fps, corresponding to approximately 5 seconds. This gives 153,600 paired frame correspondences, or 307,200 individual frames across both streams. Frames within a clip are temporally correlated, however, so 153,600 must not be interpreted as the number of independent examples; the benchmark has 1,024 clip-level source pairs and motion trajectories. The planned release manifest will include paired-clip identifiers, split membership, frame indices, demographic tags, the transfer-route label, and quality-control status. Once released with a version and integrity record, this provenance would make scoring on the fixed package repeatable even when a third-party construction backend cannot be regenerated bit-for-bit.

\subsection{Splits and Statistics}

DeepVFR adopts an identity-disjoint evaluation protocol to reduce identity leakage between training and testing. The split assigns 885 clips to training and 139 clips to testing, for 1,024 clips in total, and within-benchmark experiments follow this protocol. The external-checkpoint diagnostic in Sec.~\ref{subsec:Performance Evaluation Across Datasets} is the stated exception: because no model is fitted on DeepVFR, it scores the complete fixed collection. Identity-disjoint testing reduces memorization but does not by itself measure synthetic-to-real generalization.

The dataset is organized with appearance-based stratification tags for age group and perceived gender presentation, including child, youth, and elderly groups crossed with the retained male/female labels. These tags are provided for analysis rather than as prediction targets or claims about a person's self-identified gender. They allow future work to report not only overall performance but also subgroup behavior across different facial appearances. At the same time, the current distribution is not perfectly balanced. In particular, young-child samples are relatively fewer because current image-to-video generation and face-transfer pipelines are less stable on very young faces. This imbalance should be considered when interpreting subgroup results.

\subsection{Benchmark and Boundaries}

DeepVFR is designed to support both image-based and video-based face retouching research. Image-based methods can use the dataset as a set of paired raw/retouched frames, while video-based methods can use temporally ordered clips to model cross-frame consistency. The benchmark therefore provides a common supervision source for comparing different retouching paradigms under the same raw/retouched target definition.

Under the identity-disjoint split, each model takes the raw stream as input and is evaluated against the synchronized proxy target. This supports paired fidelity, adjacent-frame stability, qualitative inspection, and human preference studies under a common target definition. DeepVFR is not a substitute for artist-retouched captured video: both streams inherit synthetic-image and I2V characteristics, while the target additionally inherits face-transfer and blending artifacts. The benchmark therefore does not cover uncontrolled camera motion, lighting changes, sensor noise, broad pose and occlusion patterns, or the full range of subjective editing preferences.

\section{Experimental Results and Analyses}\label{sec:experiments}

We report baseline experiments that characterize DeepVFR within its controlled scope. We first describe the datasets, methods, implementation, and metrics; then report within-benchmark results, external-checkpoint transfer into DeepVFR, a human ranking study, and qualitative cases. These experiments are diagnostic rather than evidence of real-video deployment performance.

\subsection{Experimental Setup} \label{subsec:Experimental Setup}

\subsubsection{Datasets Description}

The primary benchmark used in this section is DeepVFR. As introduced in Sec.~\ref{sec:dataset}, it contains 1,024 generated-raw/constructed-target face-video pairs, with 150 frames per clip at 30 fps. Unless otherwise specified, experiments use the identity-disjoint protocol.

For the external-to-DeepVFR diagnostic, we directly apply released weights from prior work without retraining. MRFV is unavailable and FFHQR-Seq contains geometric pseudo-sequences, so a fairness-controlled external training setup cannot be reconstructed from the public materials. Because no weights are fitted on DeepVFR in this diagnostic, we evaluate all 1,024 fixed clips. This direction is external supervision $\rightarrow$ DeepVFR; it neither tests DeepVFR-trained models on captured video nor resolves the synthetic-to-real gap.

\subsubsection{Comparison Methods}

We compare methods spanning face restoration, image face retouching, and video face retouching. RestoreFormer and GPEN are restoration-oriented baselines~\citep{Wang2022RestoreFormer,Yang2021GPEN}; RetouchFormer and RetouchGPT are image-retouching methods~\citep{Wen2024RetouchFormer,Xue2025RetouchGPT}; and VRetouchEr is a video-retouching method~\citep{Xue2024VRetouchEr}. For within-benchmark experiments, the architectures are trained on the same DeepVFR paired source when applicable; the VRetouchEr setting retains its cross-frame and imperfection-flow design.

We separately test released reference weights trained on external image pairs, instructions, or pseudo-sequences. Architecture, pretraining, checkpoint, and supervision all differ, so this diagnostic is not a strict ranking and cannot isolate a causal benefit of any training dataset.

\subsubsection{Implementation and Parameter Details}

The experiments use the authors' released implementations and configurations where available. Within-benchmark models use raw DeepVFR frames or clips as input and the synchronized constructed targets as supervision, and are evaluated on the identity-disjoint test split. The external-checkpoint diagnostic instead follows the full-collection protocol described above. This separation prevents the external transfer table from being conflated with the official train/test benchmark protocol.

For the subjective study, 17 participants rank four method-anonymized results on 100 samples from the external-checkpoint evaluation. The Wenjuanxing configuration randomized both trial order and the mapping of the four anonymous candidates to the $2\!\times\!2$ grid positions; we do not claim exact counterbalancing. The supplement gives the remaining instructions and timing.

\subsubsection{Performance Metrics}

PSNR, SSIM, and LPIPS~\citep{Zhang2018LPIPS} measure frame-level agreement with the constructed target. Higher PSNR/SSIM and lower LPIPS indicate closer pixel, structure, and feature agreement. For the output-only temporal statistics, let $\hat{Y}_{i,t}$ be prediction frame $t$, $M=\sum_i(T_i-1)$, and $\mu_c(\cdot)$ the spatial mean of RGB channel $c$:
\begin{equation}
\begin{aligned}
\mathrm{LPIPS\text{-}T}
 &= \frac{1}{M}\sum_i\sum_{t=1}^{T_i-1}
 d_{\mathrm{LPIPS}}(\hat{Y}_{i,t},\hat{Y}_{i,t+1}),\\
\mathrm{RGBFlicker}
 &= \frac{1}{3M}\sum_i\sum_{t=1}^{T_i-1}\\[-2pt]
 &\quad
 \sum_{c\in\{R,G,B\}}
 \left|\mu_c(\hat{Y}_{i,t+1})-\mu_c(\hat{Y}_{i,t})\right|.
\end{aligned}
\label{eq:temporal_metrics}
\end{equation}
Predictions are decoded as RGB, area-resized to $256\!\times\!256$, and scaled to $[0,1]$ (LPIPS input: $[-1,1]$). LPIPS uses \texttt{lpips} 0.1.4, AlexNet, version 0.1. We pool all $M=20{,}711$ pairs (139 clips, 149 pairs each); neither metric uses a reference, mask, or flow. They therefore conflate genuine motion with flicker. A frozen output has zero adjacent-frame change under both definitions; RGBFlicker can also miss local changes that preserve channel means. These statistics do not establish motion preservation or perceptual temporal quality.

\begin{table}[t]
    \centering
    \small
    \caption{Results on the official identity-disjoint DeepVFR test split. Bold values mark the best target-referenced fidelity scores. Temporal columns describe unaligned adjacent-frame change; smaller values do not establish better temporal quality.}
    \label{tab:quantitative_comparison}
    \begin{tabular}{lrrrrr}
        \hline
        Model & PSNR$\uparrow$ & SSIM$\uparrow$ & LPIPS$\downarrow$ & LPIPS-T & RGBFlicker\\
        \hline
        RestoreFormer    & 24.0368 & 0.7023 & 0.2383 & 0.010897& 0.001062\\
        GPEN             & 23.9788 & 0.7299 & 0.2283 & 0.010896& 0.000828\\
        RetouchFormer    & 27.9078 & 0.8529 & 0.0937 & 0.005648 & 0.000810\\
        RetouchGPT       & \textbf{33.7426} & 0.9505 & \textbf{0.0335}& 0.005435 & 0.000816 \\
        VRetouchEr       & 33.1804 & \textbf{0.9521} & 0.0381 & 0.003507 & 0.000642\\
        \hline
    \end{tabular}
\end{table}

\subsection{Performance Evaluation on Individual Datasets} \label{subsec:Performance Evaluation on Individual Datasets}
In this subsection, we evaluate the face video retouching performance of the comparison methods on the DeepVFR benchmark, where the training and testing sets are drawn from the same dataset under the identity-disjoint protocol. The experimental results are reported in Tab.~\ref{tab:quantitative_comparison}, and the following observations can be made.
\begin{itemize}
    \item RetouchFormer, VRetouchEr, and RetouchGPT obtain PSNR values from 27.9078 to 33.7426 and SSIM values from 0.8529 to 0.9521, while LPIPS decreases from 0.0937 to 0.0335. The protocol therefore reports distinct metric profiles under common paired supervision. These values do not by themselves establish that the constructed targets are artifact-free or representative of artist retouching.
    \item RetouchGPT has the highest PSNR and lowest LPIPS, whereas VRetouchEr has the highest SSIM and smallest observed adjacent-frame changes. These results describe target agreement and output variation, rather than an established fidelity--temporal-quality trade-off. Whole-frame fidelity includes the copied background, and neither identity preservation nor the contribution of facial retouching is isolated by these scores.
    \item RestoreFormer and GPEN score below the dedicated retouching methods on target-referenced metrics. This observation is consistent with their restoration objectives being different from the benchmark's raw-to-constructed-target mapping. It should not be read as a general ranking of restoration and retouching systems outside DeepVFR.
\end{itemize}

\subsection{External-to-DeepVFR Transfer Evaluation}\label{subsec:Performance Evaluation Across Datasets}
We next apply released external checkpoints to the complete DeepVFR collection without retraining. This is a compatibility diagnostic under mismatched supervision, not an evaluation of a DeepVFR-trained model on real video. The original method papers describe the source supervision summarized in Tab.~\ref{tab:fullscope_reference}; the models nevertheless differ in architecture, pretraining, and checkpoint selection.
\begin{table}[t]
    \centering
    \small
    \caption{External checkpoints applied to all 1,024 DeepVFR clips (153,600 frames). ``Source supervision'' follows the corresponding method paper; the comparison is not fairness-controlled.}
    \label{tab:fullscope_reference}
    \begin{tabular}{llrrr}
        \hline
        Model & Source supervision & PSNR$\uparrow$ & SSIM$\uparrow$ & LPIPS$\downarrow$ \\
        \hline
        RetouchFormer  & FFHQR image pairs & 26.5413 & 0.8087 & 0.1405 \\
        RetouchGPT  & FFHQR + instructions & 26.6554 & 0.8120 & 0.1374 \\
        VRetouchEr  & FFHQR-Seq pseudo-sequences & \textbf{35.8251} & \textbf{0.9469} & \textbf{0.0451} \\
        \hline
    \end{tabular}
\end{table}

\begin{itemize}
    \item All three external checkpoints can be scored against the DeepVFR constructed targets, but their values reflect compatibility under supervision mismatch. They do not show that DeepVFR captures real-camera statistics or that training on it transfers to real videos.
    \item The released VRetouchEr checkpoint has the best values within this external-only table, but architecture, pretraining, and supervision are confounded. Its advantage therefore cannot be attributed specifically to temporal modeling or to FFHQR-Seq.
    \item Table~\ref{tab:quantitative_comparison} uses the identity-disjoint test split, whereas this diagnostic uses all 1,024 clips. Values must not be compared numerically across the two tables. A controlled supervision study would evaluate every checkpoint on the same split and hold architecture and preprocessing fixed.
\end{itemize}

\subsection{Subjective Experimental Analysis}\label{subsec:Subjective Experimental Analysis}
Seventeen participants rank four method-anonymized candidates on 100 clips from the external-checkpoint diagnostic according to visual quality, imperfection removal, naturalness, and temporal consistency. Tab.~\ref{tab:subjective_user_study} reports the aggregated ranks; the supplementary material documents the retained protocol.

\begin{table}[t]
    \centering
    \small
    \caption{Human subjective ranking results (\%) on 100 DeepVFR study clips. Higher Rank-1 indicates stronger preference within this stimulus set.}
    \label{tab:subjective_user_study}
    \begin{adjustbox}{max width=\linewidth}
    \begin{tabular}{lrrrr}
        \hline
        Candidate & Rank-1 (\%) & Rank-2 (\%) & Rank-3 (\%) & Rank-4 (\%) \\
        \hline
        DeepVFR target & \textbf{57.82} & 13.94 & 7.18 & 21.06 \\
        VRetouchEr & 22.35 & \textbf{35.59} & 31.53 & 10.53 \\
        RetouchGPT & 11.59 & 33.88 & \textbf{42.24} & 12.29 \\
        RetouchFormer & 8.24 & 16.59 & 19.06 & \textbf{56.12} \\
        \hline
    \end{tabular}
    \end{adjustbox}
\end{table}

The target receives 57.82\% Rank-1 but also 21.06\% Rank-4, approximately its combined Rank-2/3 share, revealing a substantial low-preference tail. VRetouchEr most often ranks second, whereas RetouchFormer has the largest Rank-4 share, but architecture and external supervision are confounded. Both trial order and anonymous candidate positions were randomized. The percentages nevertheless remain descriptive of this 17-participant stimulus set rather than a population estimate. The 21.06\% Rank-4 share is a fraction of judgments, not an estimated fraction of defective clips; clip-specific artifacts and participant preferences cannot be separated from these aggregates.

\subsection{Qualitative Analyses} \label{subsec:Qualitative Analyses}
The supplementary comparisons expose local mismatch and boundary artifacts hidden by averages. Selected frames do not establish temporal consistency, identity preservation, or captured-video performance, and are not used for cross-table conclusions.

\section{Discussion}\label{sec:discussion}
\noindent\textbf{Scope, artifacts, and scalability.}
DeepVFR contains 1,024 correlated, mild-motion trajectories, not a large-scale substitute for diverse captured video. Rapid speech, large pose changes, occlusion, dynamic lighting, sensor noise, camera motion, and some age groups remain underrepresented. Automatic checks and manual screening remove conspicuous failures but cannot eliminate cascaded DeepFRR/Wan2.2/transfer artifacts; manual review is the scaling bottleneck. Akool is closed and mutable, limiting exact regeneration, although a released fixed package could remain scoreable. The current work reports neither backend-stratified sensitivity nor an artist-retouched captured-video testbed.

\noindent\textbf{Evaluation boundary.}
PSNR, SSIM, and LPIPS measure agreement with a constructed reference, not identity or aesthetic correctness. LPIPS-T and RGBFlicker measure raw adjacent-frame change and do not separate motion from flicker. External checkpoints are tested only on DeepVFR, and the randomized 17-participant study remains descriptive because of its scale and stimulus scope. Stronger validation should add ArcFace identity similarity~\citep{Deng2019ArcFace}, motion-compensated error~\citep{lai2018learning,Teed2020RAFT}, video quality~\citep{Wu2023DOVER,Zhou2024THQA}, a real artist-retouched testbed, and a larger human study with an explicit balance audit.

\section{Conclusion}\label{sec:conclusion}
DeepVFR is an initial controlled synthetic benchmark---not a new architecture---with 1,024 synchronized clip pairs, an identity-disjoint protocol, and planned construction provenance. Its baselines describe target agreement, adjacent-frame output variation, external-checkpoint compatibility, and heterogeneous human preference within this domain. Synthetic and third-party components prevent claims of real-video generalization; DeepVFR is a starting point for broader, verifiable benchmarks.

\section*{CRediT authorship contribution statement}
\textbf{Peiyuan Zeng:} Investigation, Writing -- original draft.
\textbf{Fengchuang Xing:} Supervision, Writing -- review and editing.
\textbf{Shiming Wu:} Investigation, Writing -- original draft.
\textbf{Hongwei Yue:} Supervision, Writing -- review and editing.
\textbf{Chao Zhou:} Supervision, Writing -- review and editing.

\section*{Declaration of competing interest}
Peiyuan Zeng is affiliated with Cloudbutterfly Technology.

\section*{Funding}
This work was supported by the Guangdong Provincial Key Discipline Research
Capacity Enhancement Project [grant number 2025ZDJS139]; and the 2026
Provincial-Level College Student Innovation and Entrepreneurship Training
Program [grant number S202614278026].

\section*{Data availability}
The DeepVFR paired videos, individual study rankings, and training logs will
be made publicly available upon acceptance of the manuscript. The release
will include the split manifest and construction metadata described in this
article. Repository details and access information will be added when the
materials are released.

\section*{Declaration of generative AI and AI-assisted technologies in the manuscript preparation process}
OpenAI Codex was used to assist with language editing and LaTeX formatting.
The authors retain full responsibility for the final manuscript.

\bibliographystyle{elsarticle-harv}
\bibliography{main}

@misc{datareportal2025digital,
  title        = {Digital 2025: Global Overview Report},
  author       = {Kemp, Simon},
  howpublished = {DataReportal, We Are Social, and Meltwater},
  year         = {2025},
}

@inproceedings{Xue2024VRetouchEr,
  title     = {{VRetouchEr}: Learning Cross-frame Feature Interdependence with Imperfection Flow for Face Retouching in Videos},
  author    = {Xue, Wen and Jiang, Le and Xie, Lianxin and others},
  booktitle = {Proceedings of the IEEE/CVF Conference on Computer Vision and Pattern Recognition},
  pages     = {9141--9150},
  year      = {2024},
}

@article{FRRffusion2024,
  author  = {Xing, Fengchuang and Shi, Xiaowen and Wang, Yuan-Gen and others},
  title   = {Unveiling authenticity with diffusion-based face retouching reversal},
  journal = {Applied Soft Computing},
  volume  = {174},
  pages   = {113062},
  year    = {2025},
}

@inproceedings{Li2020FaceShifter,
  title     = {Advancing High Fidelity Identity Swapping for Forgery Detection},
  author    = {Li, Lingzhi and Bao, Jianmin and Yang, Hao and others},
  booktitle = {Proceedings of the IEEE/CVF Conference on Computer Vision and Pattern Recognition},
  pages     = {5074--5083},
  year      = {2020},
}

@inproceedings{SimSwap2021,
  author    = {Chen, Renwang and Chen, Xuanhong and Ni, Bingbing and others},
  title     = {{SimSwap}: An Efficient Framework for High Fidelity Face Swapping},
  booktitle = {Proceedings of the 28th ACM International Conference on Multimedia},
  pages     = {2003--2011},
  year      = {2020},
}

@inproceedings{Nirkin2019FSGAN,
  title     = {{FSGAN}: Subject Agnostic Face Swapping and Reenactment},
  author    = {Nirkin, Yuval and Keller, Yosi and Hassner, Tal},
  booktitle = {Proceedings of the IEEE/CVF International Conference on Computer Vision},
  pages     ={7184--7193},
  year      = {2019},
}

@inproceedings{Wiles2018X2Face,
  title     = {{X2Face}: A Network for Controlling Face Generation Using Images, Audio, and Pose Codes},
  author    = {Wiles, Olivia and Koepke, A. Sophia and Zisserman, Andrew},
  booktitle = {Proceedings of the European Conference on Computer Vision},
  pages     ={670-686},
  year      = {2018},
}

@inproceedings{Thies2016Face2Face,
  title     = {{Face2Face}: Real-time Face Capture and Reenactment of RGB Videos},
  author    = {Thies, Justus and Zollh{\"o}fer, Michael and Stamminger, Marc and others},
  booktitle = {Proceedings of the IEEE Conference on Computer Vision and Pattern Recognition},
  pages     ={2387-2395},
  year      = {2016},
}

@article{Thies2019NeuralTextures,
  title     = {Deferred Neural Rendering: Image Synthesis using Neural Textures},
  author    = {Thies, Justus and Zollh{\"o}fer, Michael and Nie{\ss}ner, Matthias},
  journal   = {ACM Transactions on Graphics},
  volume={38},
  number={4},
  pages={1--12},
  year      = {2019},
}

@inproceedings{Yu2018BiSeNet,
  title     = {{BiSeNet}: Bilateral Segmentation Network for Real-time Semantic Segmentation},
  author    = {Yu, Changqian and Wang, Jingbo and Peng, Chao and others},
  booktitle = {Proceedings of the European Conference on Computer Vision},
  pages     = {325--341},
  year      = {2018},
}

@inproceedings{Deng2019ArcFace,
  title     = {{ArcFace}: Additive Angular Margin Loss for Deep Face Recognition},
  author    = {Deng, Jiankang and Guo, Jia and Xue, Niannan and Zafeiriou, Stefanos},
  booktitle = {Proceedings of the IEEE/CVF Conference on Computer Vision and Pattern Recognition},
  year      = {2019},
  pages     = {4690--4699},
}

@inproceedings{Zhang2018LPIPS,
  title     = {The Unreasonable Effectiveness of Deep Features as a Perceptual Metric},
  author    = {Zhang, Richard and Isola, Phillip and Efros, Alexei A. and others},
  booktitle = {Proceedings of the IEEE Conference on Computer Vision and Pattern Recognition},
  pages     ={586--595},
  year      = {2018},
}

@inproceedings{FaceForensicsPP2019,
  author    = {R{\"o}ssler, Andreas and Cozzolino, Davide and Verdoliva, Luisa and others},
  title     = {{FaceForensics++}: Learning to Detect Manipulated Facial Images},
  booktitle = {Proceedings of the IEEE/CVF International Conference on Computer Vision},
  pages     = {1--11},
  year      = {2019},
}

@inproceedings{CelebVHQ2022,
  author    = {Zhu, Hao and Wu, Wayne and Zhu, Wentao and others},
  title     = {{CelebV-HQ}: A Large-Scale Video Facial Attributes Dataset},
  booktitle = {Proceedings of the European Conference on Computer Vision},
  pages      = {650--667},
  year      = {2022},
}

@inproceedings{SCRFD2021,
  author    = {Guo, Jia and Deng, Jiankang and Lattas, Alexandros and others},
  title     = {Sample and Computation Redistribution for Efficient Face Detection},
  booktitle = {The Tenth International Conference on Learning Representations},
  year      = {2022},
}

@inproceedings{Deng2020RetinaFace,
  title     = {{RetinaFace}: Single-Shot Multi-Level Face Localisation in the Wild},
  author    = {Deng, Jiankang and Guo, Jia and Ververas, Evangelos and others},
  booktitle = {Proceedings of the IEEE/CVF Conference on Computer Vision and Pattern Recognition},
  pages     = {5203--5212},
  year      = {2020},
}

@inproceedings{Karras2018StyleGAN,
  title     = {A Style-Based Generator Architecture for Generative Adversarial Networks},
  author    = {Karras, Tero and Laine, Samuli and Aila, Timo},
  booktitle = {Proceedings of the IEEE/CVF Conference on Computer Vision and Pattern Recognition},
  pages     = {4401--4410},
  year      = {2019},
}

@inproceedings{Teed2020RAFT,
  title     = {{RAFT}: Recurrent All-Pairs Field Transforms for Optical Flow},
  author    = {Teed, Zachary and Deng, Jia},
  booktitle = {Proceedings of the European Conference on Computer Vision},
  pages     = {402--419},
  year      = {2020},
}

@article{Kim2018DeepVideoPortraits,
  title     = {Deep Video Portraits},
  author    = {Kim, Hyeongwoo and Garrido, Pablo and Tewari, Ayush and others},
  journal   = {ACM Transactions on Graphics},
  volume={37},
  number={4},
  pages={1--14},
  year={2018},
}

@misc{TongyiWanxiang22I2VFlash,
  author       = {{Alibaba Cloud}},
  title        = {{Wan} Image-to-Video API Reference},
  howpublished = {Official product documentation},
  year         = {2026},
  note         = {Accessed: 2026-05-04},
    url          = {https://www.alibabacloud.com/help/en/model-studio/image-to-video-api-reference/}
}

@misc{AkoolAPI2025,
  author       = {{Akool}},
  title        = {{Akool} Face Swap API Overview},
  howpublished = {Official API documentation},
  year         = {2026},
  note         = {Accessed: 2026-05-04},
  url          = {https://docs.akool.com/ai-tools-suite/faceswap}
}

@misc{Wan2Repo2025,
  author       = {{Wan Team}},
  title        = {{Wan2.2}: Open and Advanced Large-Scale Video Generative Models},
  howpublished = {Official GitHub repository},
  year         = {2025},
  note         = {Accessed: 2026-05-04},
  url          = {https://github.com/Wan-Video/Wan2.2}
}

@misc{VRetouchErRepo2024,
  author       = {Xue, Wen and Jiang, Le and Xie, Lianxin and others},
  title        = {Official implementation of {VRetouchEr}},
  howpublished = {GitHub repository},
  year         = {2024},
  note         = {Accessed: 2026-05-04},
  url          = {https://github.com/Davidcoach/VRetouchEr_CVPR_2024}
}

@article{wan2025,
  title         = {Wan: Open and Advanced Large-Scale Video Generative Models},
  author        = {{Wan Team} and Wang, Ang and Ai, Baole and others},
  journal       = {arXiv preprint arXiv:2503.20314},
  year          = {2025},
  eprint        = {2503.20314},
  archivePrefix = {arXiv},
  primaryClass  = {cs.CV},
}

@inproceedings{bychkovsky2011learning,
  title     = {Learning Photographic Global Tonal Adjustment with a Database of Input / Output Image Pairs},
  author    = {Bychkovsky, Vladimir and Paris, Sylvain and Chan, Eric and others},
  booktitle = {Proceedings of the IEEE Conference on Computer Vision and Pattern Recognition},
  pages     = {97--104},
  year      = {2011},
}

@inproceedings{liang2021ppr10k,
  title     = {{PPR10K}: A Large-Scale Portrait Photo Retouching Dataset with Human-Region Mask and Group-Level Consistency},
  author    = {Liang, Jie and Zeng, Hui and Cui, Miaomiao and others},
  booktitle = {Proceedings of the IEEE/CVF Conference on Computer Vision and Pattern Recognition},
  pages     = {653--661},
  year      = {2021},
}

@inproceedings{xie2015scut,
  title     = {{SCUT-FBP}: A Benchmark Dataset for Facial Beauty Perception},
  author    = {Xie, Duorui and Liang, Lingyu and Jin, Lianwen and others},
  booktitle = {IEEE International Conference on Systems, Man, and Cybernetics},
  pages     = {1821--1826},
  year      = {2015},
  address   = {Hong Kong, China},
}

@inproceedings{liang2018scut,
  title     = {{SCUT-FBP5500}: A Diverse Benchmark Dataset for Multi-Paradigm Facial Beauty Prediction},
  author    = {Liang, Lingyu and Lin, Luojun and Jin, Lianwen and others},
  booktitle = {Proceedings of the International Conference on Pattern Recognition},
  pages     = {1598--1603},
  year      = {2018},
}

@inproceedings{lai2018learning,
  title     = {Learning Blind Video Temporal Consistency},
  author    = {Lai, Wei-Sheng and Huang, Jia-Bin and Wang, Oliver and others},
  booktitle = {Proceedings of the European Conference on Computer Vision},
  pages     ={170--185},
  year      = {2018},
}

@article{yang2022vtoonify,
  title     = {{VToonify}: Controllable High-Resolution Portrait Video Style Transfer},
  author    = {Yang, Shuai and Jiang, Liming and Liu, Ziwei and others},
  journal   = {ACM Transactions on Graphics},
  volume    = {41},
  number    = {6},
  articleno = {203},
  pages     = {1--15},
  year      = {2022},
}

@article{gebru2021datasheets,
  title   = {Datasheets for Datasets},
  author  = {Gebru, Timnit and Morgenstern, Jamie and Vecchione, Briana and others},
  journal = {Communications of the ACM},
  volume  = {64},
  number  = {12},
  pages   = {86--92},
  year    = {2021},
}

@inproceedings{ying2023retouchingffhq,
  title     = {{RetouchingFFHQ}: A Large-scale Dataset for Fine-grained Face Retouching Detection},
  author    = {Ying, Qichao and Liu, Jiaxin and Li, Sheng and others},
  booktitle = {Proceedings of the 31st ACM International Conference on Multimedia},
  pages     = {737--746},
  year      = {2023},
}

@article{lin2019exemplar,
  title   = {Exemplar-based Freckle Retouching and Skin Tone Adjustment},
  author  = {Lin, Tsung-Ying and Tsai, Yu-Ting and Huang, Tsung-Shian and others},
  journal = {Computers \& Graphics},
  volume  = {78},
  pages   = {54--63},
  year    = {2019},
}

@inproceedings{bae2023digiface,
  title     = {{DigiFace-1M}: 1 Million Digital Face Images for Face Recognition},
  author    = {Bae, Gwangbin and de La Gorce, Martin and Baltru{\v{s}}aitis, Tadas and others},
  booktitle = {Proceedings of the IEEE/CVF Winter Conference on Applications of Computer Vision},
  pages     = {3526--3535},
  year      = {2023},
}

@inproceedings{shafaei2021autoretouch,
  title     = {{AutoRetouch}: Automatic Professional Face Retouching},
  author    = {Shafaei, Alireza and Little, James J. and Schmidt, Mark},
  booktitle = {Proceedings of the IEEE/CVF Winter Conference on Applications of Computer Vision},
  pages     = {990--998},
  year      = {2021},
}

@inproceedings{xu2025face,
  title     = {Face Retouching with Diffusion Data Generation and Spectral Restorement},
  author    = {Xu, Zhidan and Zhang, Xiaoqin and Lu, Shijian},
  booktitle = {Proceedings of the IEEE/CVF International Conference on Computer Vision},
  pages     = {14722--14731},
  year      = {2025},
}

@inproceedings{Zhu2021MegaFS,
  author    = {Zhu, Yuhao and Li, Qi and Wang, Jian and others},
  title     = {One Shot Face Swapping on Megapixels},
  booktitle = {Proceedings of the IEEE/CVF Conference on Computer Vision and Pattern Recognition},
  pages     = {4834--4844},
  year      = {2021},
}

@inproceedings{Wang2021HifiFace,
  author    = {Wang, Yuhan and Chen, Xu and Zhu, Junwei and others},
  title     = {{HifiFace}: 3D Shape and Semantic Prior Guided High Fidelity Face Swapping},
  booktitle = {Proceedings of the Thirtieth International Joint Conference on Artificial Intelligence},
  pages     = {1136--1142},
  year      = {2021},
}

@inproceedings{Xu2022HighResolutionFaceSwapping,
  author    = {Xu, Yangyang and Deng, Bailin and Wang, Junle and others},
  title     = {High-Resolution Face Swapping via Latent Semantics Disentanglement},
  booktitle = {Proceedings of the IEEE/CVF Conference on Computer Vision and Pattern Recognition},
  pages     = {7642--7651},
  year      = {2022},
}

@inproceedings{Rosberg2023FaceDancer,
  author    = {Rosberg, Felix and Aksoy, Eren Erdal and Alonso-Fernandez, Fernando and others},
  title     = {{FaceDancer}: Pose- and Occlusion-Aware High Fidelity Face Swapping},
  booktitle = {Proceedings of the IEEE/CVF Winter Conference on Applications of Computer Vision},
  pages     = {3454--3463},
  year      = {2023},
}

@inproceedings{Li2023ThreeDAwareFaceSwapping,
  author    = {Li, Yixuan and Ma, Chao and Yan, Yichao and others},
  title     = {3D-Aware Face Swapping},
  booktitle = {Proceedings of the IEEE/CVF Conference on Computer Vision and Pattern Recognition},
  pages     = {12705--12714},
  year      = {2023},
}

@inproceedings{Shiohara2023BlendFace,
  author    = {Shiohara, Kaede and Yang, Xingchao and Taketomi, Takafumi},
  title     = {{BlendFace}: Re-designing Identity Encoders for Face-Swapping},
  booktitle = {Proceedings of the IEEE/CVF International Conference on Computer Vision},
  pages     = {7634--7644},
  year      = {2023},
}

@inproceedings{StyleGAN22020,
  author    = {Karras, Tero and Laine, Samuli and Aittala, Miika and others},
  title     = {Analyzing and Improving the Image Quality of StyleGAN},
  booktitle = {Proceedings of the IEEE/CVF Conference on Computer Vision and Pattern Recognition},
  pages     = {8107--8116},
  year      = {2020},
}

@article{Wen2024RetouchFormer,
  author  = {Wen, Xue and Xie, Lianxin and Jiang, Le and Chen, Tianyi and Wu, Si and Liu, Cheng and Wong, Hau-San},
  title   = {{RetouchFormer}: Semi-supervised High-Quality Face Retouching Transformer with Prior-Based Selective Self-Attention},
  journal = {Proceedings of the AAAI Conference on Artificial Intelligence},
  volume  = {38},
  number  = {6},
  pages   = {5903--5911},
  year    = {2024},
  doi     = {10.1609/aaai.v38i6.28404},
  url     = {https://ojs.aaai.org/index.php/AAAI/article/view/28404},
}

@article{Xue2025RetouchGPT,
  author  = {Xue, Wen and Ding, Chun and Xu, Ruotao and Wu, Si and Xu, Yong and Wong, Hau-San},
  title   = {{RetouchGPT}: {LLM}-based Interactive High-Fidelity Face Retouching via Imperfection Prompting},
  journal = {Proceedings of the AAAI Conference on Artificial Intelligence},
  volume  = {39},
  number  = {9},
  pages   = {9059--9067},
  year    = {2025},
  doi     = {10.1609/aaai.v39i9.32980},
  url     = {https://ojs.aaai.org/index.php/AAAI/article/view/32980},
}

@inproceedings{Wang2022RestoreFormer,
  author    = {Wang, Zhouxia and Zhang, Jiawei and Chen, Runjian and Wang, Wenping and Luo, Ping},
  title     = {{RestoreFormer}: High-Quality Blind Face Restoration From Undegraded Key-Value Pairs},
  booktitle = {Proceedings of the IEEE/CVF Conference on Computer Vision and Pattern Recognition},
  pages     = {17512--17521},
  year      = {2022},
  url       = {https://openaccess.thecvf.com/content/CVPR2022/html/Wang_RestoreFormer_High-Quality_Blind_Face_Restoration_From_Undegraded_Key-Value_Pairs_CVPR_2022_paper.html},
}

@inproceedings{Yang2021GPEN,
  author    = {Yang, Tao and Ren, Peiran and Xie, Xuansong and Zhang, Lei},
  title     = {{GAN} Prior Embedded Network for Blind Face Restoration in the Wild},
  booktitle = {Proceedings of the IEEE/CVF Conference on Computer Vision and Pattern Recognition},
  pages     = {672--681},
  year      = {2021},
  url       = {https://openaccess.thecvf.com/content/CVPR2021/html/Yang_GAN_Prior_Embedded_Network_for_Blind_Face_Restoration_in_the_CVPR_2021_paper.html},
}

@inproceedings{Yan2024DF40,
  author    = {Yan, Zhiyuan and Yao, Taiping and Chen, Shen and Zhao, Yandan and Fu, Xinghe and Zhu, Junwei and Luo, Donghao and Wang, Chengjie and Ding, Shouhong and Wu, Yunsheng and Yuan, Li},
  title     = {{DF40}: Toward Next-Generation Deepfake Detection},
  booktitle = {Advances in Neural Information Processing Systems},
  volume    = {37},
  pages     = {29387--29434},
  publisher = {Curran Associates, Inc.},
  year      = {2024},
  doi       = {10.52202/079017-0925},
  url       = {https://proceedings.neurips.cc/paper_files/paper/2024/hash/34239f60eca7ce9bee5280aaf81362d8-Abstract-Datasets_and_Benchmarks_Track.html},
}

@inproceedings{Khalid2021FakeAVCeleb,
  author    = {Khalid, Hasam and Tariq, Shahroz and Kim, Minha and Woo, Simon S.},
  title     = {{FakeAVCeleb}: A Novel Audio-Video Multimodal Deepfake Dataset},
  booktitle = {Thirty-fifth Conference on Neural Information Processing Systems Datasets and Benchmarks Track},
  year      = {2021},
  url       = {https://openreview.net/forum?id=TAXFsg6ZaOl},
}

@inproceedings{Liu2025MoFRR,
  author    = {Liu, Jiaxin and Ying, Qichao and Qian, Zhenxing and Li, Sheng and Zhang, Runqi and Liu, Jian and Zhang, Xinpeng},
  title     = {{MoFRR}: Mixture of Diffusion Models for Face Retouching Restoration},
  booktitle = {Proceedings of the IEEE/CVF International Conference on Computer Vision},
  pages     = {12842--12851},
  year      = {2025},
  url       = {https://openaccess.thecvf.com/content/ICCV2025/html/Liu_MoFRR_Mixture_of_Diffusion_Models_for_Face_Retouching_Restoration_ICCV_2025_paper.html},
}

@inproceedings{Wu2026RetouchIQ,
  author    = {Wu, Qiucheng and Shi, Jing and Jenni, Simon and Kafle, Kushal and Wang, Tianyu and Chang, Shiyu and Zhao, Handong},
  title     = {{RetouchIQ}: {MLLM} Agents for Instruction-Based Image Retouching with Generalist Reward},
  booktitle = {Proceedings of the IEEE/CVF Conference on Computer Vision and Pattern Recognition},
  pages     = {12279--12288},
  year      = {2026},
  url       = {https://openaccess.thecvf.com/content/CVPR2026/html/Wu_RetouchIQ_MLLM_Agents_for_Instruction-Based_Image_Retouching_with_Generalist_Reward_CVPR_2026_paper.html},
}

@inproceedings{Yang2026BeautyGRPO,
  author    = {Yang, Jiachen and Lin, Xianhui and Dong, Yi and Zheng, Zebiao and Liu, Xing and Gu, Hong and Fang, Yanmei},
  title     = {{BeautyGRPO}: Aesthetic Alignment for Face Retouching via Dynamic Path Guidance and Fine-Grained Preference Modeling},
  booktitle = {Proceedings of the IEEE/CVF Conference on Computer Vision and Pattern Recognition},
  pages     = {25110--25120},
  year      = {2026},
  url       = {https://openaccess.thecvf.com/content/CVPR2026/papers/Yang_BeautyGRPO_Aesthetic_Alignment_for_Face_Retouching_via_Dynamic_Path_Guidance_CVPR_2026_paper.pdf},
}

@inproceedings{Wu2023DOVER,
  author    = {Wu, Haoning and Zhang, Erli and Liao, Liang and Chen, Chaofeng and Hou, Jingwen and Wang, Annan and Sun, Wenxiu and Yan, Qiong and Lin, Weisi},
  title     = {Exploring Video Quality Assessment on User Generated Contents from Aesthetic and Technical Perspectives},
  booktitle = {Proceedings of the IEEE/CVF International Conference on Computer Vision},
  pages     = {20144--20154},
  year      = {2023},
  url       = {https://openaccess.thecvf.com/content/ICCV2023/html/Wu_Exploring_Video_Quality_Assessment_on_User_Generated_Contents_from_Aesthetic_ICCV_2023_paper.html},
}

@inproceedings{Zhou2024THQA,
  author    = {Zhou, Yingjie and Zhang, Zicheng and Sun, Wei and Liu, Xiaohong and Min, Xiongkuo and Wang, Zhihua and Zhang, Xiao-Ping and Zhai, Guangtao},
  title     = {{THQA}: A Perceptual Quality Assessment Database for Talking Heads},
  booktitle = {2024 IEEE International Conference on Image Processing},
  pages     = {15--21},
  year      = {2024},
  doi       = {10.1109/ICIP51287.2024.10647507},
  url       = {https://cmsworkshops.com/ICIP2024/view_paper.php?PaperNum=1437},
}

@inproceedings{Garrido2014Reenactment,
  author    = {Garrido, Pablo and Valgaerts, Levi and Rehmsen, Ole and Thormahlen, Thorsten and Perez, Patrick and Theobalt, Christian},
  title     = {Automatic Face Reenactment},
  booktitle = {Proceedings of the IEEE Conference on Computer Vision and Pattern Recognition},
  pages     = {4217--4224},
  year      = {2014},
  url       = {https://openaccess.thecvf.com/content_cvpr_2014/html/Garrido_Automatic_Face_Reenactment_2014_CVPR_paper.html},
}
\end{document}


\begin{frontmatter}
\title{Supplementary Material: From Retouched Images to Paired Videos: A Synthetic Benchmark for Face Video Retouching}
\renewcommand{\ead}[1]{}
\renewcommand{\corref}[1]{}
\renewcommand{\cortext}[2][]{}
\author[aff1,aff2]{Peiyuan Zeng}
\ead{andyzeng0921@outlook.com}
\author[aff1]{Fengchuang Xing\corref{cor1}}
\ead{xingfengchuang@gdei.edu.cn}
\author[aff1]{Shiming Wu}
\ead{hotgirllllllll@outlook.com}
\author[aff1]{Hongwei Yue}
\ead{springfall@foxmail.com}
\author[aff3]{Chao Zhou}
\ead{zhouchaodzkd@e.gzhu.edu.cn}
\cortext[cor1]{Corresponding author.}
\affiliation[aff1]{organization={College of Physics and Information Engineering,
  Guangdong University of Education},country={China}}
\affiliation[aff2]{organization={Cloudbutterfly Technology}}
\affiliation[aff3]{organization={School of Electronic Information, Huzhou College},
  city={Huzhou},postcode={313000},country={China}}

\end{frontmatter}
\renewcommand{\thesection}{S\arabic{section}}
\renewcommand{\thetable}{S\arabic{table}}
\renewcommand{\thefigure}{S\arabic{figure}}
\section{Additional Documentation}

This supplement documents construction settings, the retained human-study
protocol, and qualitative cases for experiments already reported in the main
paper. It does not introduce results on additional datasets.

\subsection{Construction Details}
DeepVFR is constructed from paired raw/retouched synthetic face images through a two-stage pipeline. First, a source portrait clip is synthesized from the raw image using Tongyi Wanxiang 2.2 I2V with a mild motion prompt. Second, the retouched appearance is transferred to the generated clip, with Akool as the primary production backend and SimSwap used for a smaller complementary subset. An automatic detection/alignment validity screen and subsequent visual inspection filter conspicuous failures before the pairs enter the retained collection.

For construction transparency, Table~\ref{tab:appendix_construction_settings} summarizes the settings that should accompany the versioned package planned for release upon acceptance. Once released, these records could support repeatable use of fixed clips; they do not imply bit-exact regeneration of results produced by a mutable commercial API.

\begin{table}[!htbp]
\centering
\caption{Key settings used in the DeepVFR construction pipeline.}
\label{tab:appendix_construction_settings}
\begin{tabularx}{\textwidth}{@{}p{0.27\textwidth}X@{}}
\hline
Item & Setting \\
\hline
Source supervision & Paired raw/retouched face images from DeepFRR \\
Video synthesis model & Tongyi Wanxiang 2.2 I2V \\
Motion prompt & Mild facial-motion prompt (e.g., smile / micro-expression) \\
Clip length & 150 frames per video \\
Frame rate & 30 fps \\
Duration & About 5 seconds per clip \\
Primary transfer route & Akool-based face transfer \\
Secondary transfer route & SimSwap-based complementary subset \\
Output pair format & Temporally aligned raw/retouched video pairs \\
Automatic validity screen & Successful face detection and backend-specific alignment \\
Visual quality screen & Remove motion collapse, obvious boundary artifacts, retouch-style loss, and clear identity drift \\
Per-clip provenance & Pair identifier, frame index, transfer-route label, split, demographic tag, and QC status \\
Release protocol & Identity-disjoint split for within-benchmark experiments; full collection for the no-retraining external diagnostic \\
Regeneration boundary & Akool service updates may prevent bit-exact reconstruction; released fixed clips would remain scoreable \\
\hline
\end{tabularx}
\end{table}

\begin{figure}[!htbp]
    \centering
    \begin{minipage}[t]{0.42\textwidth}
        \centering
        \includegraphics[width=\textwidth]{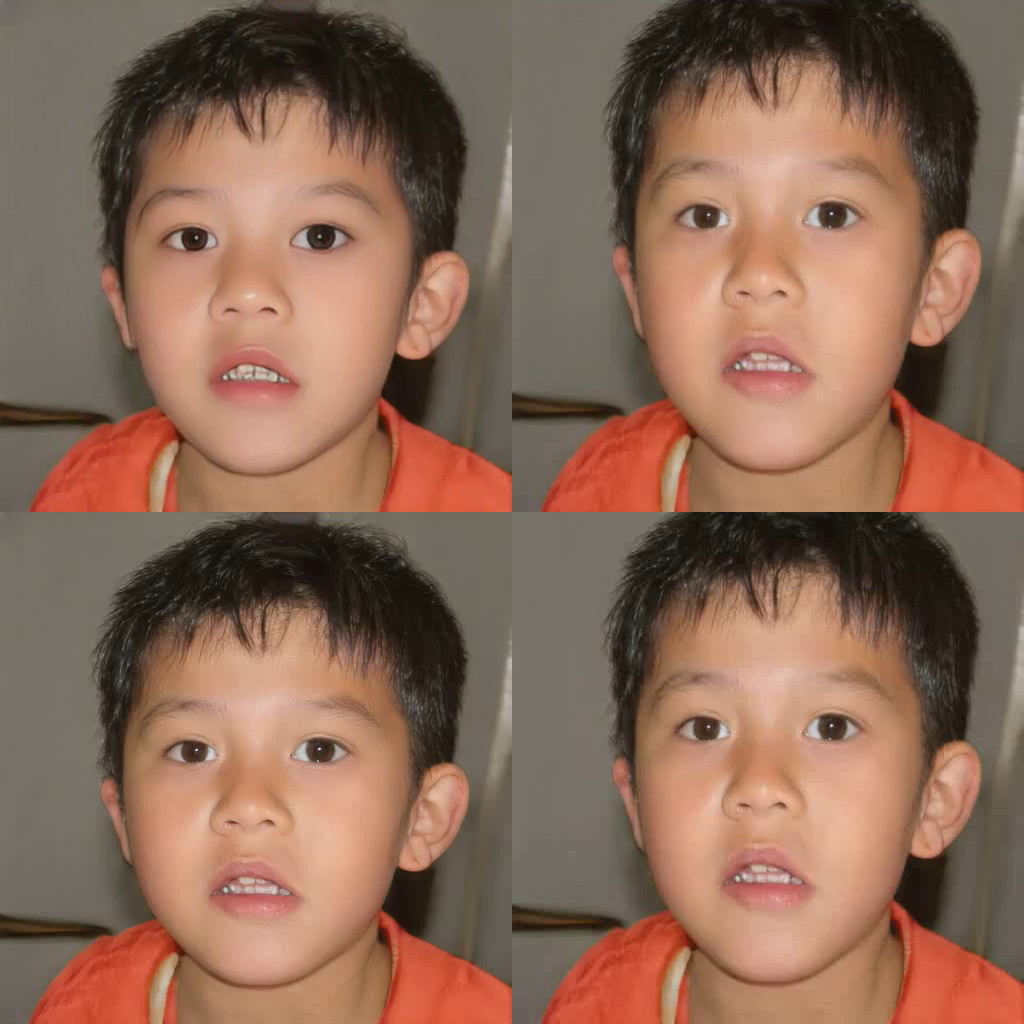}
    \end{minipage}
    \hfill
    \begin{minipage}[t]{0.42\textwidth}
        \centering
        \includegraphics[width=\textwidth]{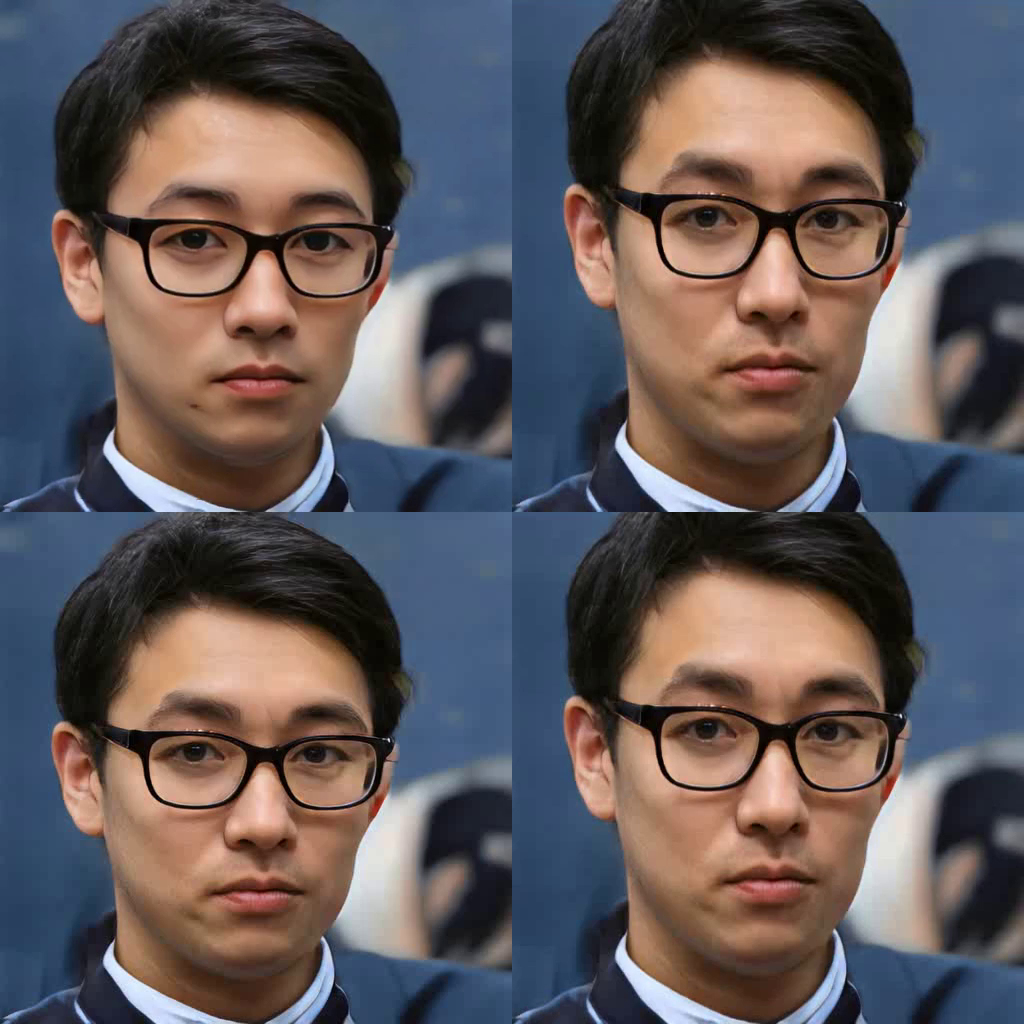}
    \end{minipage}
    \caption{Representative first-frame screenshots from two grid videos used in the human study. Each clip presents four method-anonymized candidates for direct comparison.}
    \label{fig:appendix_grid_trials}
\end{figure}

Auxiliary preprocessing organizes paired clips, extracts frames for inspection, and converts assets to the input format expected by each evaluated model. The per-clip transfer route must be included in the planned release manifest because the current paper does not report a backend-stratified sensitivity analysis. Representative human-study grid videos are shown in Fig.~\ref{fig:appendix_grid_trials}.

\subsection{Human Subjective Study Protocol}
The subjective study was conducted in a controlled remote setting. Before the formal evaluation, participants received unified instructions through Tencent Meeting, including the study goal, ranking criteria, and operating procedure. During the evaluation, each trial displayed one indexed grid video containing four method-anonymized retouching results. After a clip finished playing, the system waited 15 seconds before switching to the next trial. Each study section was repeated three times, and the final rankings were collected through Wenjuanxing for aggregation and validity checking. The Wenjuanxing configuration randomized both trial order and the assignment of the anonymous candidates to the four grid positions.

Table~\ref{tab:appendix_user_study_settings} summarizes the main protocol settings used in the human study. Participants were asked to rank the compared results according to overall visual quality, imperfection removal, naturalness, and temporal stability. The questionnaire backend additionally recorded metadata such as submission time and dwell time, which were used only for anonymous response validation. The video-presentation format is illustrated by the grid examples in Fig.~\ref{fig:appendix_grid_trials}.

\begin{table}[!htbp]
\centering
\caption{Main settings of the human subjective study.}
\label{tab:appendix_user_study_settings}
\small
\begin{tabularx}{\textwidth}{@{}p{0.27\textwidth}X@{}}
\hline
Item & Setting \\
\hline
Participants & 17 \\
Evaluated samples & 100 video clips \\
Compared results per trial & 4 anonymous retouching results shown simultaneously \\
Viewing format & Grid video playback with indexed trial prompts \\
Rating rule & Rank the four compared results from best to worst \\
Evaluation criteria & Overall visual quality, imperfection removal, naturalness, and temporal stability \\
Briefing stage & Unified instructions delivered through Tencent Meeting \\
Response collection & Wenjuanxing questionnaire \\
Trial interval & 15 seconds before switching to the next video \\
Section repetition & 3 times per section \\
Validation metadata & Submission time and dwell time \\
Randomization & Trial order and anonymous candidate-to-grid-position mapping \\
\hline
\end{tabularx}
\end{table}

These supplementary notes clarify the practical construction route of DeepVFR and the execution protocol of the human study.

\subsection{Qualitative Comparisons and Failure Cases}

\begin{figure}[!htbp]
    \centering
    \includegraphics[width=0.86\textwidth]{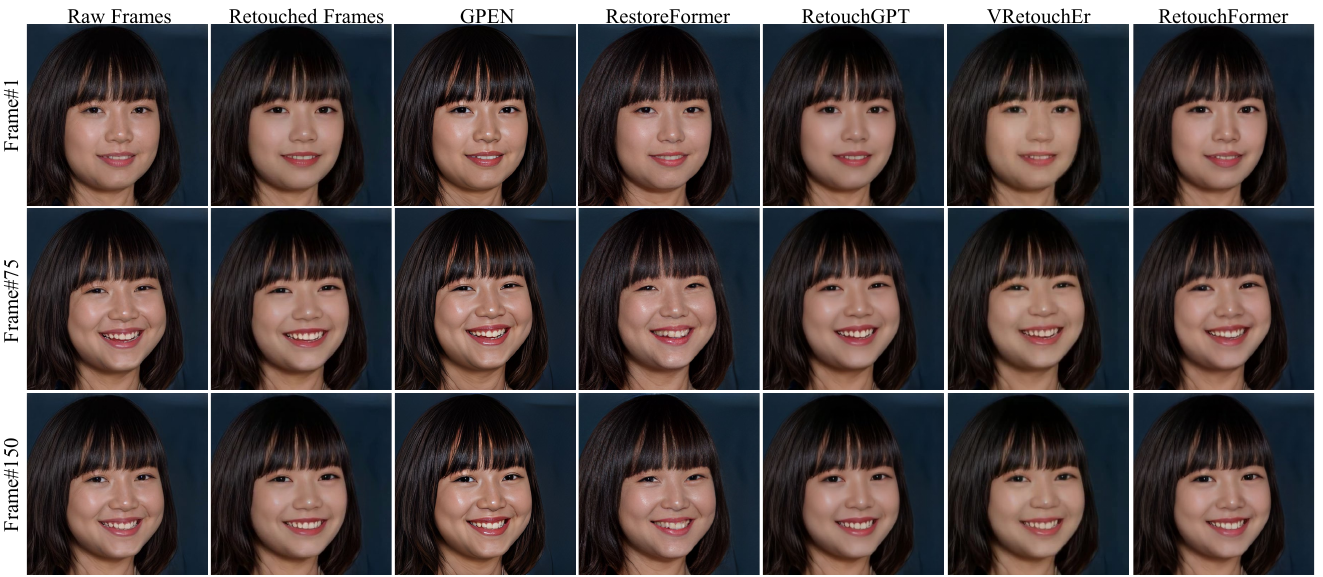}
    \caption{Qualitative comparison on frames from one evaluated clip. From left to right: raw input, constructed proxy target (labeled ``Retouched Frames'' in the visualization), GPEN, RestoreFormer, RetouchGPT, VRetouchEr, and RetouchFormer. In this example, the dedicated retouching methods more closely follow the constructed target than the restoration baselines.}
    \label{fig:supp_qualitative_comparison}
\end{figure}

Figure~\ref{fig:supp_qualitative_comparison} compares methods on nearby frames from one clip. Restoration-oriented baselines deviate from the constructed target in local appearance and skin-refinement strength, consistent with their different task objective. Dedicated retouching methods are visually closer to the target in these examples. The nearby frames permit qualitative inspection, but a few selected frames cannot establish temporal consistency on their own.

\begin{figure}[!htbp]
    \centering
    \includegraphics[width=0.86\textwidth]{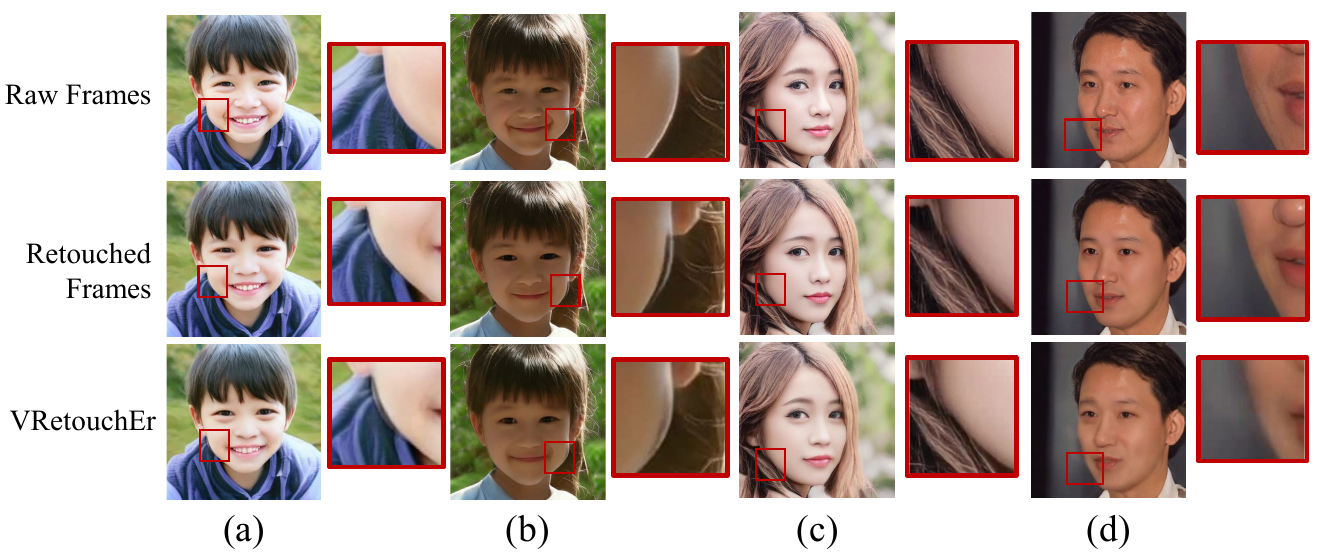}
    \caption{Representative successful and challenging cases of VRetouchEr trained on DeepVFR. Rows correspond to raw frames, constructed proxy-target frames (labeled ``Retouched Frames'' in the visualization), and VRetouchEr outputs. Examples (a) and (b) show relatively successful results, while (c) and (d) show local deviations from the proxy targets. Red boxes and zoomed regions highlight local differences.}
    \label{fig:supp_vretoucher_case_study}
\end{figure}

In Fig.~\ref{fig:supp_vretoucher_case_study}, Examples (a) and (b) follow the moderate constructed edit while retaining expression and surrounding content visually. Examples (c) and (d) retain deviations around facial regions and blend boundaries. These cases are consistent with difficulty in subtle skin refinement, local color agreement, and boundary handling. We do not infer identity preservation because no face-recognition metric was collected. The examples remain within the controlled synthetic domain and do not demonstrate performance under real camera motion or artist-authored retouching.